\documentclass[letterpaper, 10 pt,journal, twoside]{IEEEtran}  

\IEEEoverridecommandlockouts                              

\usepackage{graphics} 
\usepackage{epsfig} 
\usepackage{times} 
\usepackage{amsmath} 
\usepackage{amssymb}  
\usepackage{booktabs} 
\usepackage{amsfonts}
\usepackage[colorlinks,linkcolor=black,anchorcolor=black,citecolor=black,urlcolor=black]{hyperref}
\usepackage[table]{xcolor}
\usepackage{booktabs}
\usepackage{multirow}
\usepackage{graphicx}
\usepackage{subfigure}
\usepackage{cite}
\usepackage{float}
\usepackage{stfloats}
\usepackage{cuted}
\usepackage{capt-of}
\usepackage{makecell}
\usepackage{adjustbox}
\usepackage[ruled,vlined,linesnumbered]{algorithm2e}
\usepackage{multirow}
\usepackage{booktabs}
\SetArgSty{textnormal} 
\makeatletter
\renewcommand{\@algocf@pre@ruled}{%
  \kern 1.5mm 
  \hrule height\algoheightrule depth0pt
  \kern\interspacetitleruled
}
\makeatother

\makeatletter
\let\IEEEdefaultmakecaption\@makecaption
\long\def\@makecaption#1#2{%
  \ifx\@captype\@IEEEtablestring
    \footnotesize\bgroup\par\centering\@IEEEtabletopskipstrut
    {\normalfont\footnotesize #1.\ }{\normalfont\footnotesize #2}%
    \par\addvspace{0.5\baselineskip}\egroup%
    \@IEEEtablecaptionsepspace
  \else
    \IEEEdefaultmakecaption{#1}{#2}%
  \fi
}
\def\tablename{Table}
\makeatother

\graphicspath{{../Fig/}{../}{Fig/}{img/}}

\title{
Observation Quality Matters: Robust Multi-Fisheye Calibration via Failure-Oriented Analysis
}

\author{Peize Liu$^{1}$, Zhe Tong$^{1}$, Chen Feng$^{1,\dag}$, and Shaojie Shen$^{1}$
\thanks{$^{1}$Department of Electronic and Computer Engineering, The Hong Kong University of Science and Technology, Hong Kong, China.}
\thanks{Email: $\{$pliuan, ztongae, cfengag, eeshaojie$\}$@ust.hk}
\thanks{\textbf{$^{\dag}$Corresponding Author}}
}

\begin{document}

\bstctlcite{IEEEexample:BSTcontrol}
\pagestyle{empty}

\makeatletter
\def\@IEEENORMtitlevspace{0.5\baselineskip}
\def\@IEEEMINtitlevspace{0.5\baselineskip}
\makeatother
\maketitle
\thispagestyle{empty}

\begin{abstract}
Reliable calibration of multi-fisheye camera systems remains challenging as rig size, camera arrangement diversity, and field of view increase. 
Existing pipelines can jointly optimize intrinsics, extrinsics, and target poses, but their success still depends heavily on empirical capture rules and the quality of the observations supplied to the solver. 
This paper studies this dependency through a failure-oriented analysis. 
We reveal that calibration failures are not sufficiently explained by detector recall loss or global image-plane distribution imbalance. 
Instead, the dominant failure factor lies in intrinsic initialization: observations with limited radial span couple focal scale with fisheye projection-shape parameters, producing ill-conditioned updates. 
Guided by this insight, we propose CO-Calib, a plug-in calibration-data construction framework that combines a robust learning-based target detector with an error-analysis-guided frame selector. 
CO-Calib constructs initialization-friendly anchors, co-visible multi-camera constraints, and coverage-completion frames without changing the existing calibration workflow or optimization backend. 
Extensive experiments on synthetic and real multi-fisheye systems demonstrate that CO-Calib improves the overall success rate from 68.1\% to 99.3\%, increases extrinsic accuracy, and augments real-world calibration stability.
The source code will be made publicly available at \url{https://github.com/HKUST-Aerial-Robotics/CO-Calib}.
\end{abstract}


\vspace{-0.4cm}
\section{Introduction}
\label{sec:intro}

\IEEEPARstart{M}{ulti-fisheye} camera systems equipped with large-field-of-view (FoV) lenses are widely used in mobile robots \cite{mavic_3t,skydio_x10} and data collection platforms \cite{yogamani2019woodscape, perfetti2026ant3d}, where they improve perception performance in tasks such as state estimation \cite{qin2017vins} and depth prediction \cite{OmniMVS,Xie_OmniVidar}. 
Yet robust calibration of such systems remains challenging. 
As the number of cameras, their spatial arrangement, and the covered FoV increase, the underlying bundle-adjustment (BA) problem becomes larger and more tightly coupled. 
Meanwhile, accurate calibration requires reliable observations over a broader and more distorted image domain, making the optimization increasingly sensitive to the quality and distribution of the observations supplied to the solver.

This observation sensitivity is particularly evident in industrial workflows. 
Different rig configurations are often calibrated with tailored procedures that prescribe capture trajectories, per-camera sampling requirements, and sometimes even dedicated calibration targets such as 3D or spherical patterns. 
As a result, seemingly minor changes in this process can induce noticeably different outcomes or calibration failure, leading to repeated engineering effort and limited robustness across platforms.

Existing multi-fisheye calibration methods, such as Kalibr \cite{extend_kalibr}, typically follow a representative pipeline that progressively initializes and refines camera intrinsics, per-view target poses, and inter-camera extrinsics. 
Subsequent studies improve calibration from different perspectives, for example by boosting intrinsic estimation under severe distortion \cite{duisterhof2022tartancalib} or by strengthening geometric constraints through richer calibration targets \cite{mc-calib}. 
Taken together, these works reveal that calibration performance is governed not only by the nonlinear BA solver, but depends critically on the observations on which the solver operates.
However, observation quality is still treated largely in an implicit and empirical manner. 
What remains insufficiently understood is a fundamental question: under what observational conditions does a multi-fisheye calibration problem become well-conditioned and optimization-ready?

To answer this research question, we conduct a systematic failure-oriented analysis of existing calibration pipelines and identify a central finding: calibration failure is not determined simply by detector recall or global image-plane distribution. 
The decisive factor is whether the observations make intrinsic initialization well conditioned. 
In multi-fisheye calibration, observations with limited radial span can couple focal scale with fisheye projection-shape parameters, producing unstable linearized updates. 
This is the sense in which \emph{observation quality matters}: useful observations must provide separable parameter directions, not merely more detected target points or more uniform image-plane coverage. 
The quantitative analysis is presented in Sec.~\ref{sec:failure_analysis}.

Building on this analysis, we propose a simple yet principled solution that explicitly improves observation quality throughout the calibration process.
Benefiting from the strong function approximation capability of neural networks, we devise a learning-based calibration target detector that enables accurate and robust detection under severe distortion and appearance variations.
We further design an error-analysis-guided frame selector that organizes the detected observations into an initialization-friendly sequence.
The selector first chooses anchor frames with broad radial coverage to stabilize intrinsic initialization, then preserves high-quality co-visible multi-camera constraints and fills the remaining weakly constrained image regions for later refinement.
Rather than redesigning existing calibration pipelines, this study takes an analysis-driven perspective to understand why multi-fisheye calibration fails and how observation properties shape optimization behavior, yielding a plug-in approach that substantially enhances robustness.

Experimental results on both synthetic and real-world multi-fisheye systems demonstrate that the proposed analysis-guided strategies substantially improve calibration success rate and accuracy. 
On the synthetic benchmark, the proposed method increases the overall calibration success rate from 68.1\% to 99.3\%, while also improving extrinsic calibration accuracy, particularly under large-FoV configurations. 
On real-world datasets, our method achieves robust performance across different stereo configurations and significantly improves calibration stability on the challenging Hex-Fisheye system.

Our contributions are summarized as follows:
\begin{enumerate}
\item We present a failure-oriented analysis of multi-fisheye calibration and show that initialization failures are not explained by detection recall or global image-plane distribution alone, but are strongly associated with ill-conditioned intrinsic updates caused by insufficient parameter separability.
\item We identify the focal--projection coupling as a key observation-dependent failure mechanism in fisheye initialization, and show that observations with broad radial coverage are critical for separating focal scale from projection-shape parameters.
\item We propose \textbf{CO-Calib}, a plug-in calibration-data construction framework that combines a robust learning-based target detector with an error-analysis-guided frame selector to provide reliable detections, stable initialization anchors, co-visible multi-camera constraints, and coverage completion for refinement.
\item We validate the proposed analysis and framework on both controlled synthetic and real-world multi-fisheye systems, demonstrating improved calibration robustness and accuracy compared to existing pipelines. 
The source code of our implementation will be released at \href{https://github.com/HKUST-Aerial-Robotics/CO-Calib}{\textcolor{red}{{https://github.com/HKUST-Aerial-Robotics/CO-Calib}}}.
\end{enumerate}

\vspace{-0.4cm}
\section{Related Work}
\label{sec:related_works}

\subsection{Fisheye Camera Model and Calibration}
\label{sec:camera_models}

Large-FoV fisheye imaging is nontrivial because its strongly nonlinear projection exceeds the capability of conventional perspective models.
Existing large-FoV camera models address this issue mainly in three ways.
One line of work formulates the projection through a unified geometric representation of viewing rays, as in omnidirectional models such as Mei's unified model and later geometric central models \cite{mei_model,double_sphere_model}.
Other studies \cite{hughes2010equidistant,kannala_brandt_model} model the fisheye projection more directly by expressing the image radius as a function of the incident ray angle. 
A third line of work \cite{Radial_polynomial_omnidirectional} uses generic polynomial approximations of the projection or inverse projection function. 
These models provide practical formulations for large-FoV cameras, including highly distorted fisheye systems.

Despite these advances in camera modeling, accurate fisheye calibration remains challenging since the model parameters must be reliably constrained by observations across the full image domain. 
In practice, observations in heavily distorted peripheral regions are often sparse or low-quality, making accurate model parameterization difficult. 
TartanCalib \cite{duisterhof2022tartancalib} addresses this issue by predicting undetected calibration targets using an intermediate camera model and then refining the model with the expanded set of target observations. 
While effective in improving peripheral coverage, such approaches remain sensitive to the quality of the intermediate model, and prediction errors may propagate into the subsequent optimization and degrade parameter estimation accuracy.
Motivated by this limitation, our work tackles this issue at the source by introducing a robust learning-based target detector.

\vspace{-0.4cm}
\subsection{Multi-Fisheye System Calibration}
\label{sec:calib_pipelines}
Multi-fisheye calibration is commonly formulated as a BA-based nonlinear least-squares problem. 
Kalibr has become a widely used toolbox built on this formulation \cite{unified_multi_sensor,extend_kalibr}. 
Its pipeline progressively activates the calibration variables: camera intrinsics are first initialized, then refined together with per-view poses, and finally optimized jointly with the inter-camera extrinsics. 
This fully coupled strategy can achieve high accuracy when initialization and observations are reliable, but its robustness often degrades as the number of cameras and optimization variables increases, making the enlarged joint problem hard to solve stably.
To improve robustness in more challenging configurations, later works introduce richer calibration targets and more structured observation formation. 
For example, MC-Calib \cite{mc-calib} increases geometric constraints via multi-target design and staged estimation, showing that calibration performance can be enhanced by enriching the observation structure presented to the optimizer.

Although their procedures differ, Kalibr and MC-Calib are ultimately solved within a similar optimization framework. 
Their practical distinction lies less in the optimizer itself than in how observations and constraints are formed before and during optimization. 
These methods suggest that enriching calibration observations can improve calibration performance, yet how such observations affect optimization remains insufficiently characterized.
Particularly, it remains unclear what kinds of observations support both stable initialization and accurate final estimation in multi-fisheye calibration. 
Addressing this gap is the central goal of our work.

\vspace{-0.3cm}
\section{Failure Mechanism Analysis}
\label{sec:failure_analysis}

All analyses in this section are conducted using controlled calibration datasets. To systematically study multi-fisheye calibration, we simplify the camera configurations into four representative stereo arrangements with relative rotations of $0^\circ$, $60^\circ$, $90^\circ$, and $120^\circ$. Based on these four arrangements, we further evaluate calibration performance under four FoV levels, namely $180^\circ$, $200^\circ$, $220^\circ$, and $240^\circ$, resulting in a total of 16 experimental settings. For each setting, we generate 100 calibration sequences, each containing 480 synchronized frames. 
Each sequence is treated as one calibration trial. 
A trial is considered successful if the pipeline terminates normally and returns valid calibration parameters with bounded reprojection errors; otherwise, it is counted as a failure.
Detailed dataset generation and experimental settings are provided in Sec.~\ref{sec:experiments}.

\vspace{-0.4cm}
\subsection{Preliminary: Problem Formulation and Pipeline}
Multi-fisheye calibration is formulated as a joint nonlinear least-squares problem over known calibration-target points \(\mathbf{P}_j \in \mathbb{R}^3\). For camera \(i\) and calibration view \(k\), let \({}^{i}\mathbf{u}^{k}_{j} \in \mathbb{R}^2\) denote the detected image point, and let \(\Omega\) collect all valid camera-view-point observations.

The optimization variables include the intrinsic parameters \(\boldsymbol{\theta}_i\), the rig-to-camera extrinsics \(\mathbf{T}^{c_i}_{r} \in SE(3)\), and the per-view target poses \({}^{r}\mathbf{T}^{k} \in SE(3)\) in the rig frame:
{
\begin{equation}
\mathcal{X}
=
\left(
\{\boldsymbol{\theta}_i\}_{i=0}^{M-1},
\{\mathbf{T}^{c_i}_{r}\}_{i=0}^{M-1},
\{{}^{r}\mathbf{T}^{k}\}_{k=1}^{K}
\right),
\label{eq:optimization_state}
\end{equation}
}
where \(M\) is the number of cameras and \(K\) is the number of calibration views.
Given \(\mathcal{X}\), the predicted measurement of point \(\mathbf{P}_j\) is
{
\begin{equation}
{}^{i}\hat{\mathbf{u}}^{k}_{j}
=
\pi_i\!\left(
\mathbf{T}^{c_i}_{r}\,{}^{r}\mathbf{T}^{k}\mathbf{P}_j;\,
\boldsymbol{\theta}_i
\right),
\label{eq:predicted_measurement}
\end{equation}
}
where \(\pi_i(\cdot;\boldsymbol{\theta}_i)\) is the fisheye projection model. The reprojection residual is
{
\begin{equation}
{}^{i}\mathbf{r}^{k}_{j}(\mathcal{X})
=
{}^{i}\mathbf{u}^{k}_{j}
-
{}^{i}\hat{\mathbf{u}}^{k}_{j},
\label{eq:reprojection_residual}
\end{equation}
}
and calibration minimizes the robust weighted reprojection error
{
\begin{equation}
\min_{\mathcal{X}}
\sum_{(i,k,j)\in\Omega}
\rho\!\left(
{}^{i}\mathbf{r}^{k}_{j}(\mathcal{X})^{\top}
\mathbf{\Sigma}_{ikj}^{-1}
{}^{i}\mathbf{r}^{k}_{j}(\mathcal{X})
\right),
\label{eq:calibration_objective}
\end{equation}
}
where \(\mathbf{\Sigma}_{ikj}\) is the measurement covariance and \(\rho(\cdot)\) is a robust kernel.

This formulation couples intrinsics and extrinsics through shared target poses: observations from the same view share \({}^{r}\mathbf{T}^{k}\), while each camera observes that pose through its rig-to-camera transform. In practice, the problem is solved by a staged pipeline:
1) \textbf{Intrinsic Initialization}, which progressively estimates camera intrinsics and per-view target poses;
2) \textbf{Intrinsic Refinement}, which refines target poses and full intrinsic parameters; and
3) \textbf{Full Calibration}, which jointly optimizes inter-camera extrinsics, intrinsics, and target poses.

\vspace{-0.4cm}
\subsection{Localization of Calibration Failures}

\vspace{-0.5cm}
\begin{table}[H]
  \centering
  \caption{Failure localization across FoV and stereo configurations. Each entry reports the calibration failure rate / the fraction of failed trials attributed to intrinsic initialization failure.}
  \vspace{-0.3cm}
  \label{tab:failure_localization}
  \setlength{\tabcolsep}{3pt}
  \renewcommand{\arraystretch}{0.95}
  \scriptsize
  \begin{tabular*}{\columnwidth}{@{\extracolsep{\fill}}lcccc}
    \toprule
    \textbf{FoV}
    & \multicolumn{4}{c}{\textbf{Stereo configuration}} \\
    \cmidrule(lr){2-5}
    & \textbf{0$^\circ$}
    & \textbf{60$^\circ$}
    & \textbf{90$^\circ$}
    & \textbf{120$^\circ$} \\
    \midrule
    180$^\circ$ & 3.0\% / 0.0\% & 0.0\% / -- & 0.0\% / -- & 0.0\% / -- \\
    200$^\circ$ & 3.0\% / 100.0\% & 2.0\% / 100.0\% & 3.0\% / 100.0\% & 8.0\% / 75.0\% \\
    220$^\circ$ & 67.0\% / 98.5\% & 36.0\% / 100.0\% & 44.0\% / 100.0\% & 46.0\% / 100.0\% \\
    240$^\circ$ & 88.0\% / 98.9\% & 59.0\% / 100.0\% & 75.0\% / 100.0\% & 77.0\% / 100.0\% \\
    \bottomrule
  \end{tabular*}
  \vspace{-0.3cm}
\end{table}

We first localize failures within the staged calibration pipeline. 
Table \ref{tab:failure_localization} reports, for each FoV and stereo configuration, the overall calibration failure rate and the fraction of failed trials caused by intrinsic-initialization failure. Failures are rare at narrower FoVs but increase sharply for wide-FoV settings; when calibration fails, almost all failed trials originate from intrinsic initialization.

This localization narrows the analysis to intrinsic initialization, where intrinsics and per-view target poses are estimated from calibration-target observations. We evaluate three observation-related hypotheses:
1) \textbf{Insufficient peripheral constraints:} severe fisheye distortion may reduce target recall near the image boundary, leaving peripheral regions weakly constrained.
2) \textbf{Imbalanced spatial distribution:} non-parallel camera arrangements may skew image-plane observations and bias intrinsic estimation.
3) \textbf{Ill-conditioned intrinsic initialization:} 
The strong nonlinearity of the fisheye camera model can make the focal scale and projection-shape parameters difficult to disentangle during initialization, leading to an ill-conditioned optimization problem that is fragile to solve.

The following parts test these hypotheses and show that degraded recall and spatial imbalance are secondary; the dominant factor is poor separability among intrinsic parameter directions during initialization.

\vspace{-0.5cm}
\subsection{Insufficient Peripheral Constraints}
\label{sec:recall_constraint_analysis}

\vspace{-0.4cm}
\begin{table}[h]
  \centering
  \caption{Calibration success rate using geometry-based detections and full ground-truth observations. Each cell reports geometry-based/GT success rate.}
  \vspace{-0.3cm}
  \label{tab:gt_observation_effect}
  \setlength{\tabcolsep}{3pt}
  \renewcommand{\arraystretch}{0.9}
  \scriptsize
  \begin{tabular*}{\columnwidth}{@{\extracolsep{\fill}}lcccc}
    \toprule
    \textbf{FoV}
    & \multicolumn{4}{c}{\textbf{Stereo configuration}} \\
    \cmidrule(lr){2-5}
    & \textbf{0$^\circ$}
    & \textbf{60$^\circ$}
    & \textbf{90$^\circ$}
    & \textbf{120$^\circ$} \\
    \midrule
    180$^\circ$ & 97\%$\,$/$\,$100\% & 100\%$\,$/$\,$99\% & 100\%$\,$/$\,$97\% & 100\%$\,$/$\,$96\% \\
    200$^\circ$ & 97\%$\,$/$\,$78\% & 98\%$\,$/$\,$97\% & 97\%$\,$/$\,$82\% & 92\%$\,$/$\,$78\% \\
    220$^\circ$ & 33\%$\,$/$\,$9\% & 64\%$\,$/$\,$34\% & 56\%$\,$/$\,$35\% & 54\%$\,$/$\,$24\% \\
    240$^\circ$ & 12\%$\,$/$\,$2\% & 41\%$\,$/$\,$14\% & 25\%$\,$/$\,$9\% & 23\%$\,$/$\,$5\% \\
    \midrule
    \textbf{Total} & \multicolumn{4}{c}{68.1\%$\,$/$\,$53.7\%} \\
    \bottomrule
  \end{tabular*}
  \vspace{-0.2cm}
\end{table}

\begin{figure}[t]
\centering
\includegraphics[width=0.99\columnwidth]{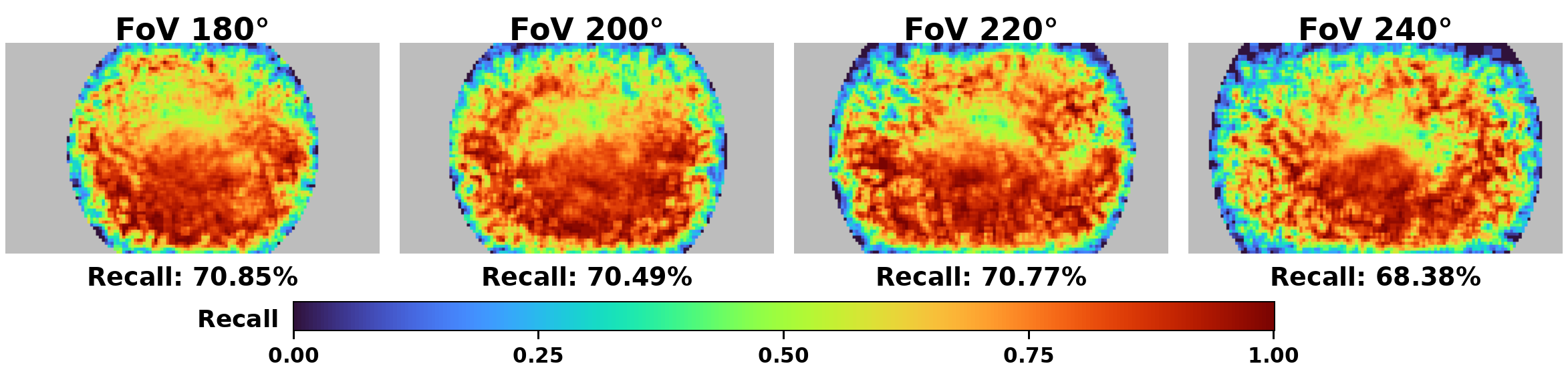}
\vspace{-0.7cm}
\caption{Recall degradation of the geometry-based detector in fisheye images. The recall rate decreases as the FoV increases.}
\label{fig:coverage_degradation}
\vspace{-0.5cm}
\end{figure}

Target detection in fisheye images is particularly vulnerable to severe distortion, especially near the image boundary. The first hypothesis is therefore that calibration fails because reduced peripheral recall leaves too few observations to constrain the high-distortion part of the camera model. Sparse or rejected boundary observations weaken the initialization stage precisely where fisheye parameters are most sensitive. Fig.~\ref{fig:coverage_degradation} confirms this effect: recall decreases near the image boundary, making peripheral regions less constrained than central regions. 

To test whether this missing-constraint effect is the dominant cause, we replace detected target points with full ground-truth observations and rerun the same calibration pipeline. 
If insufficient peripheral constraints mainly induced the failure, ground-truth observations should improve calibration success. 
Table \ref{tab:gt_observation_effect} shows the opposite: the success rate drops from 68.1\% to 53.7\%, and most wide-FoV settings become less stable. Thus, calibration failure cannot be explained by low recall or insufficient peripheral sample count alone; full observations can instead expose the underlying optimization instability.

\vspace{-0.5cm}
\subsection{Imbalanced Spatial Distribution}
\label{sec:distribution_analysis}

The second hypothesis is that calibration failure is driven by imbalanced image-plane constraints. We test this hypothesis at two levels: distribution shifts across stereo configurations and distribution differences between successful and failed trials within the same configuration. If spatial imbalance were the dominant factor, the failure pattern should align with one of these two distribution changes.

\begin{figure}[t]
\centering
\hspace*{0.3cm}
\includegraphics[width=0.9\columnwidth]{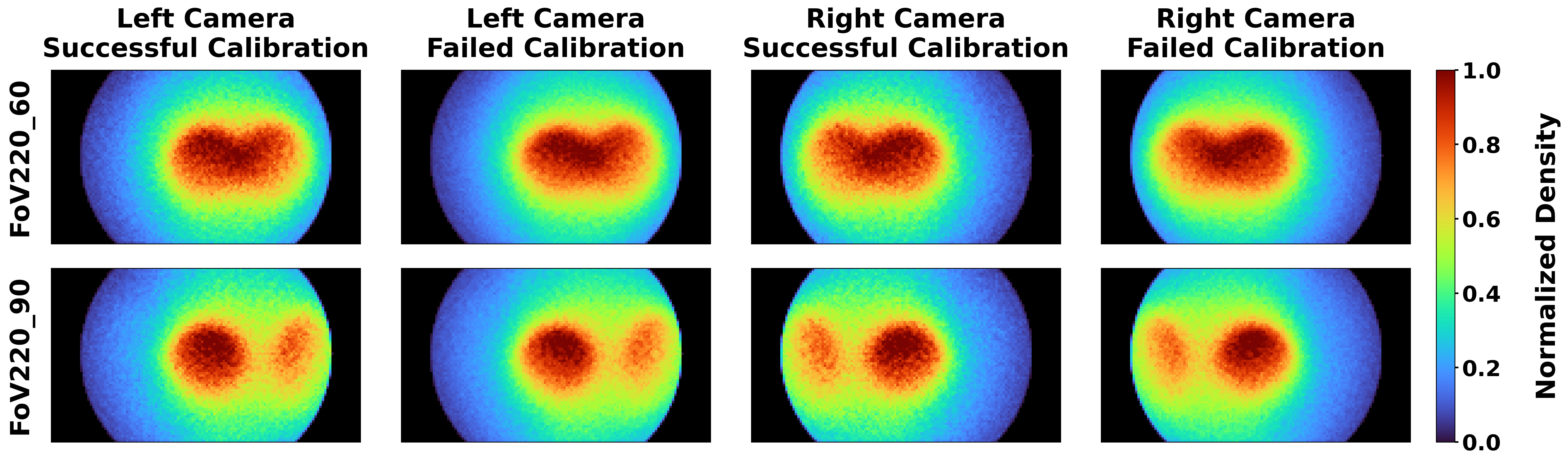}
\vspace{-0.3cm}
\caption{Spatial constraint distributions across configurations and calibration outcomes. FoV220\_60 and FoV220\_90 exhibit different global distributions, while successful and failed cases within the same configuration show similar image-plane patterns.}
\label{fig:distribution_cross_success_failure}
\vspace{-0.2cm}
\end{figure}

Fig.~\ref{fig:distribution_cross_success_failure} shows that FoV220\_60 and FoV220\_90 have visibly different observation distributions because the stereo arrangement changes where the target projects onto the image plane. However, the two settings have similar success rates under GT observations, suggesting that a configuration-level distribution shift alone does not explain the failures. Within each configuration, successful and failed trials also exhibit similar GT observation patterns. We quantify this comparison using the observation-count difference \(D_{\mathrm{qty}}\), the spatial-distribution distance \(D_{\mathrm{sp}}\), the random-label baseline \(D_{\mathrm{rand}}\), and \(\Delta_{\mathrm{sp}} = |D_{\mathrm{sp}} - D_{\mathrm{rand}}|\), with detailed definitions in Appendix~\ref{app:spatial_distribution_metrics}.

\begin{table}[h]
  \centering
  \caption{Success/failure distribution comparison within each configuration.}
  \vspace{-0.3cm}
  \label{tab:distribution_success_failure_gt_raw_nohuber}
  \setlength{\tabcolsep}{2.0pt}
  \renewcommand{\arraystretch}{0.86}
  \scriptsize
  \begin{tabular*}{\columnwidth}{@{\extracolsep{\fill}}lcccc}
    \toprule
    \textbf{Setting} 
    & \textbf{$D_{\mathrm{qty}}$} 
    & \textbf{$D_{\mathrm{sp}}$} 
    & \textbf{$D_{\mathrm{rand}}$}
    & \textbf{$\Delta_{\mathrm{sp}}$} \\
    \midrule
    FoV200\_0   & 0.15\% & 3.00\%  & 2.99\%  & 0.01 pp \\
    FoV200\_60  & 0.15\% & 3.94\%  & 4.00\%  & 0.06 pp \\
    FoV200\_90  & 0.29\% & 3.58\%  & 3.62\%  & 0.04 pp \\
    FoV200\_120 & 0.11\% & 3.64\%  & 3.68\%  & 0.04 pp \\
    FoV220\_0   & 0.01\% & 4.64\%  & 4.65\%  & 0.01 pp \\
    FoV220\_60  & 0.12\% & 2.86\%  & 2.92\%  & 0.06 pp \\
    FoV220\_90  & 0.28\% & 2.97\%  & 3.05\%  & 0.08 pp \\
    FoV220\_120 & 0.22\% & 3.78\%  & 3.68\%  & 0.10 pp \\
    FoV240\_0   & 0.05\% & 10.25\% & 10.12\% & 0.13 pp \\
    FoV240\_60  & 0.13\% & 4.20\%  & 4.20\%  & 0.00 pp \\
    FoV240\_90  & 0.25\% & 5.21\%  & 5.33\%  & 0.12 pp \\
    FoV240\_120 & 0.29\% & 7.55\%  & 7.46\%  & 0.09 pp \\
    \bottomrule
  \end{tabular*}
  \vspace{-0.5cm}
\end{table}

Table~\ref{tab:distribution_success_failure_gt_raw_nohuber} confirms that the point-count difference is negligible: \(D_{\mathrm{qty}}\) remains below 0.30\% in all settings, with a median of 0.15\% and a mean of 0.17\%. Moreover, \(D_{\mathrm{sp}}\) closely matches \(D_{\mathrm{rand}}\), and \(\Delta_{\mathrm{sp}}\) is at most 0.13 percentage points. Thus, the success/failure split is no more spatially separated than a random split, indicating that neither observation count nor image-plane distribution is sufficient to explain the initialization failures.

\vspace{-0.4cm}
\subsection{Ill-Conditioned Intrinsic Initialization}
\label{sec:condition_analysis}

After excluding detector recall and spatial distribution as primary causes, we examine whether failed cases induce a more ill-conditioned intrinsic-initialization problem. 
During intrinsic initialization, frames are progressively incorporated into the optimization rather than optimized only once using the full sequence. 
Let \(K\) denote the total number of frames used by the initialization procedure, and let
\(\mathcal{I}_k=\{1,\ldots,k\}\), \(k\le K\), denote the frame prefix used at the \(k\)-th initialization step. 
At this step, the optimizer linearizes the problem around the current estimate
$
\bar{\mathbf{x}}^{(k)}
=
\left[
\bar{\mathbf{x}}_{c}^{(k)},
\bar{\mathbf{x}}_{p}^{(k)}
\right],
$
where \(\bar{\mathbf{x}}_{c}^{(k)}\) is the current camera-intrinsic estimate and 
\(\bar{\mathbf{x}}_{p}^{(k)}\) stacks the target-pose estimates associated with the first \(k\) frames.

Let \(\mathbf{r}^{(k)}\) be the residual vector constructed from the observations in \(\mathcal{I}_k\). 
The Jacobian evaluated at the current linearization point is
{
\begin{equation}
\mathbf{J}^{(k)}
=
\left.
\frac{\partial \mathbf{r}^{(k)}}
{\partial [\mathbf{x}_{c},\mathbf{x}_{p}]}
\right|_{\bar{\mathbf{x}}^{(k)}}
=
\begin{bmatrix}
\mathbf{J}_{c}^{(k)} & \mathbf{J}_{p}^{(k)}
\end{bmatrix},
\end{equation}
}
where \(\mathbf{J}_{c}^{(k)}\) and \(\mathbf{J}_{p}^{(k)}\) are the residual Jacobians with respect to the camera-intrinsic parameters and target poses, respectively. 
The corresponding normal equation matrix is
{
\begin{equation}
\mathbf{H}^{(k)}
=
\left(\mathbf{J}^{(k)}\right)^{\top}
\mathbf{J}^{(k)}
=
\begin{bmatrix}
\mathbf{H}_{cc}^{(k)} & \mathbf{H}_{cp}^{(k)} \\
\mathbf{H}_{pc}^{(k)} & \mathbf{H}_{pp}^{(k)}
\end{bmatrix}.
\end{equation}
}
Here, 
\(\mathbf{H}_{cc}^{(k)}
=
\left(\mathbf{J}_{c}^{(k)}\right)^{\top}
\mathbf{J}_{c}^{(k)}\)
is the direct information block for the estimated camera parameters, including focal lengths, principal points, and projection-shape parameters.

Because target poses are nuisance variables, we evaluate the camera-parameter information after pose elimination using the Schur complement:
{
\begin{equation}
\mathbf{S}_{c}^{(k)}
=
\mathbf{H}_{cc}^{(k)}
-
\mathbf{H}_{cp}^{(k)}
\left(\mathbf{H}_{pp}^{(k)}\right)^{+}
\mathbf{H}_{pc}^{(k)} ,
\end{equation}
}
where \((\cdot)^{+}\) denotes the pseudo-inverse. 
This pose-free block measures the local information available for updating the camera parameters at the \(k\)-th initialization step.

To remove the influence of different parameter units and scales, we normalize the Schur block with a diagonal scaling matrix \(\mathbf{D}^{(k)}\):
{
\begin{equation}
\tilde{\mathbf{S}}_{c}^{(k)}
=
\mathbf{D}^{(k)}
\mathbf{S}_{c}^{(k)}
\mathbf{D}^{(k)} .
\end{equation}
}
We then define the conditioning score at the \(k\)-th initialization step as
{
\begin{equation}
\gamma_k
=
\log_{10}
\kappa
\left(
\tilde{\mathbf{S}}_{c}^{(k)}
\right)
=
\log_{10}
\frac{
\max_i |\lambda_i^{(k)}|
}{
\min_i |\lambda_i^{(k)}|
},
\end{equation}
}
where \(\{\lambda_i^{(k)}\}\) are the eigenvalues of 
\(\tilde{\mathbf{S}}_{c}^{(k)}\). 
Thus, each calibration trial produces a condition-number trajectory
\(\{\gamma_k\}_{k=1}^{K}\) along the intrinsic-initialization path. 
Larger values indicate weaker local separability among camera-parameter directions after eliminating target poses.

\begin{figure}[h]
\centering
\hspace*{-0.1cm}
\includegraphics[width=1.0\columnwidth]{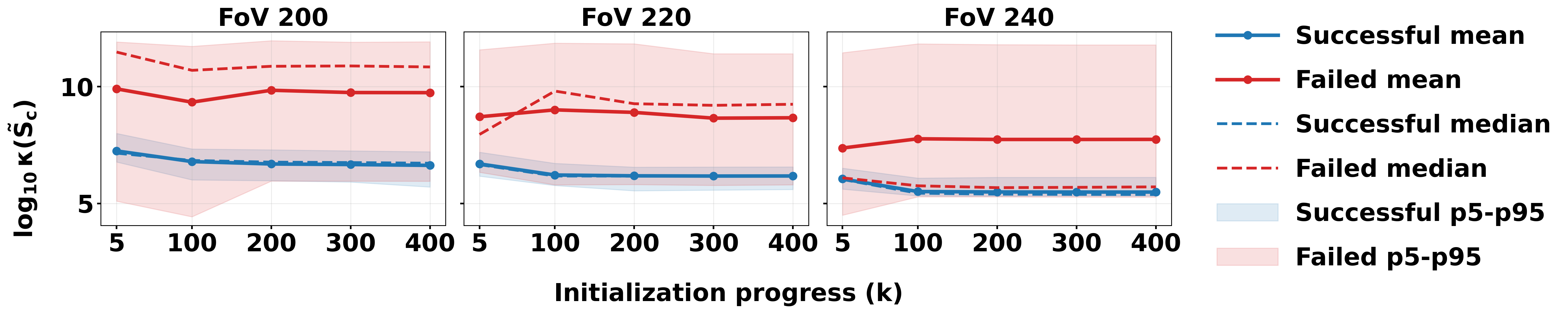}
\vspace{-0.6cm}
\caption{Condition-number distribution of the normalized pose-free camera-parameter block. Failed cases exhibit higher \(\log_{10}\kappa(\tilde{\mathbf{S}}_{c})\), indicating weaker and more ambiguous intrinsic constraints.}
\label{fig:estimated_parameter_condition_distribution}
\vspace{-0.2cm}
\end{figure}

Fig.~\ref{fig:estimated_parameter_condition_distribution} compares the conditioning scores of successful and failed trials along the initialization path. 
Failed cases consistently exhibit larger \(\gamma_k\) values than successful cases, indicating that they induce more ill-conditioned intrinsic updates at the current linearization points encountered by the optimizer. 
Therefore, the observed failures are associated with poor local parameter separability rather than only missing detections or image-plane spatial imbalance. 
In a linearized least-squares problem, such ill-conditioning means that different parameter-update directions can produce nearly indistinguishable residual changes, which suggests ambiguity between focal scale and projection-shape parameters.

Following Eq.~\eqref{eq:predicted_measurement}, the fisheye projection of a normalized ray \(\mathbf{r}\) can be locally written as
{
\begin{equation}
\hat{\mathbf{u}}
=
\pi(\mathbf{r};\boldsymbol{\theta})
=
\mathbf{F}
\boldsymbol{\phi}(\mathbf{r};\boldsymbol{\eta})
+
\mathbf{c},
\end{equation}
}
where \(\mathbf{F}=\mathrm{diag}(f_x,f_y)\), 
\(\mathbf{c}=[c_x,c_y]^\top\), and 
\(\boldsymbol{\eta}\) denotes projection-shape parameters such as \(\xi\), \(\alpha\), or \(\beta\). 
The main ambiguity lies in the term 
\(\mathbf{F}\boldsymbol{\phi}(\mathbf{r};\boldsymbol{\eta})\): 
both focal scale and projection shape can change the radial image coordinate. 
When observations occupy only a narrow radial range, their Jacobian directions can become locally similar, making these parameters difficult to separate during initialization.

To verify that this ambiguity is not implementation-specific, we analyze Omni, EUCM, and Double-Sphere projection families under different radial profiles. 
For sampled rays \(\mathcal{R}=\{\mathbf{r}_{q}\}_{q=1}^{N}\), we use a scalar focal scale \(\mathbf{F}=f\mathbf{I}\) and form the stacked projection response
{
\begin{equation}
\mathbf{y}(f,\boldsymbol{\eta};\mathcal{R})
= [f\,\boldsymbol{\phi}(\mathbf{r}_{1};\boldsymbol{\eta}) \cdots f\,\boldsymbol{\phi}(\mathbf{r}_{N};\boldsymbol{\eta})]^{\top}.
\end{equation}
}
We compute the focal--projection coupling from the normalized Jacobian:
{
\begin{equation}
\begin{aligned}
\mathbf{J}_{f\eta}
&=
\frac{\partial \mathbf{y}(f,\boldsymbol{\eta};\mathcal{R})}
{\partial [f,\boldsymbol{\eta}]},
&
\left(\hat{\mathbf{J}}_{f\eta}\right)_{:,j}
&=
\frac{
\left(\mathbf{J}_{f\eta}\right)_{:,j}
}{
\left\|
\left(\mathbf{J}_{f\eta}\right)_{:,j}
\right\|_{2}
},\\
\mathbf{G}_{f\eta}
&=
\hat{\mathbf{J}}_{f\eta}^{\top}
\hat{\mathbf{J}}_{f\eta},
&
C_{f\eta}
&=
\log_{10}
\kappa(\mathbf{G}_{f\eta}).
\end{aligned}
\label{eq:focal_projection_coupling_index}
\end{equation}
}
A larger \(C_{f\eta}\) indicates stronger linear dependence among focal-scale and projection-shape Jacobian directions.

\begin{figure}[t]
\centering
\includegraphics[width=0.9\columnwidth]{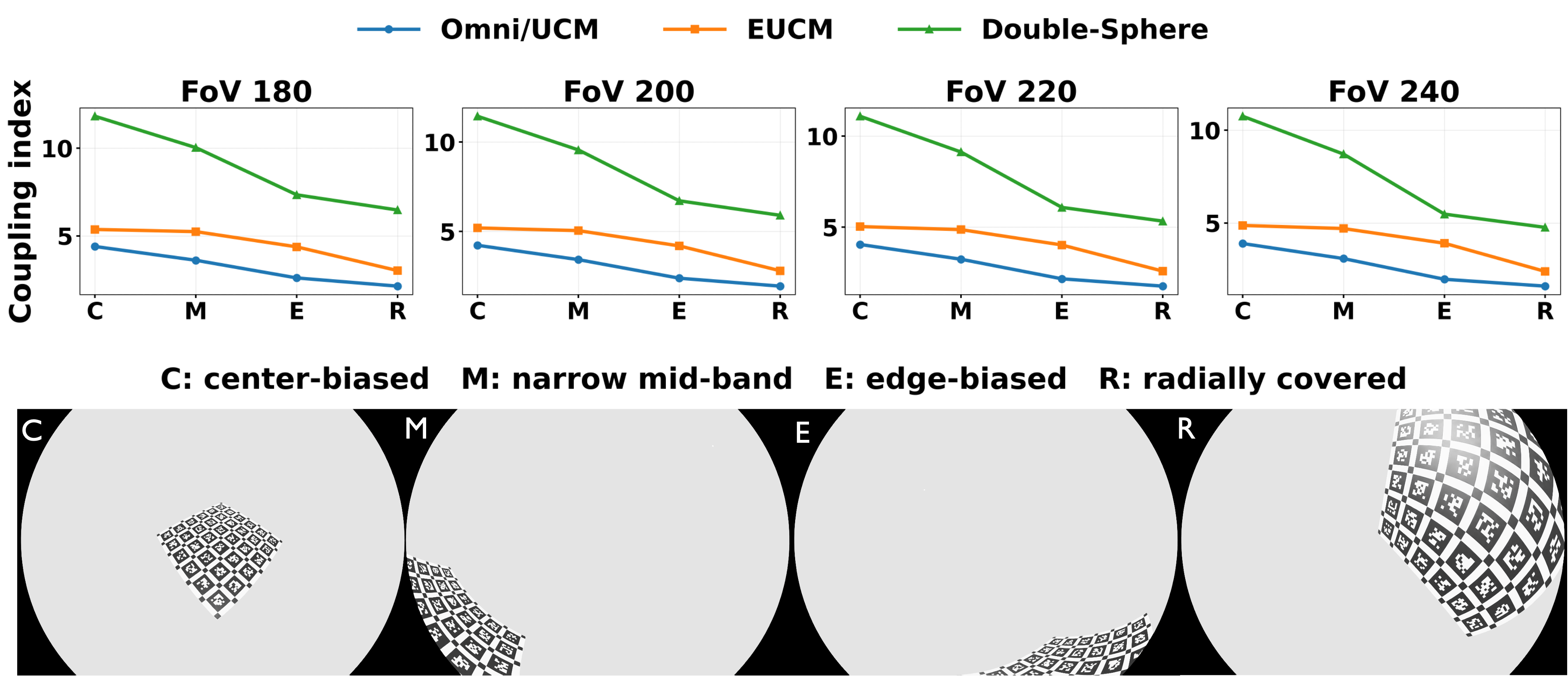}
\vspace{-0.4cm}
\caption{Projection--focal coupling under different radial profiles. Single-band observations (C/M/E) show stronger coupling than radially covered observations (R), indicating that broad radial coverage improves projection--focal separability.}
\label{fig:focal_projection_coupling}
\vspace{-0.7cm}
\end{figure}

Fig.~\ref{fig:focal_projection_coupling} compares central (C), middle-radius (M), edge (E), and radially covered (R) profiles. 
Single-band profiles consistently produce stronger coupling, while radial coverage yields the lowest coupling across all three fisheye families. 
This confirms that focal--projection coupling is a general property of fisheye projection Jacobians under insufficient radial coverage, and that stable intrinsic initialization requires observations spanning a broad radial range.

\vspace{-0.5cm}
\subsection{Key Findings}
\label{key_findings}

This analysis demonstrates that low recall and spatial imbalance are observable in multi-fisheye calibration, but neither is the primary failure mechanism. 
The core issue is the poor separability of intrinsic parameter directions during initialization.
Unlike pinhole calibration, fisheye calibration introduces stronger coupling between focal scale and projection-shape parameters, and when observations occupy only a narrow radial range, these parameters can produce similar residual changes and become difficult to estimate reliably.
Stable calibration, thus, requires observations that actively decouple the fisheye parameters and provide a well-conditioned sequence of linearized updates, so that initialization can progressively approach the final estimate. 
Moreover, Experiment~\ref{sec:calibration_experiment} shows that bypassing this linearized initialization path sharply reduces success rate, showing that final joint optimization cannot compensate for poorly conditioned initialization.

\vspace{-0.4cm}
\section{CO-Calib}
\label{sec:method}

\vspace{-0.2cm}
\subsection{Framework Overview}

Motivated by the failure analysis in Sec.~\ref{sec:failure_analysis}, we propose \textbf{CO-Calib}, a framework that constructs optimization-ready calibration observations to stabilize multi-fisheye calibration. CO-Calib does not modify the camera model or bundle-adjustment backend. Instead, it improves the observations supplied to the optimizer so that intrinsic initialization receives reliable detections and a well-conditioned frame sequence.

CO-Calib contains two components. First, a learning-based calibration-target detector improves recall and localization stability under severe fisheye distortion. Second, an error-analysis-guided frame selector constructs an optimization-ready frame sequence according to the identified failure mechanisms. The detector increases usable observations, while the selector chooses frames that support stable and progressive estimation of fisheye camera parameters.

\vspace{-0.4cm}
\subsection{Robust Calibration Target Detection}

CO-Calib uses a learning-based target detector trained with an online, physically grounded data-generation pipeline. Unlike generic methods, it is designed specifically for calibration targets and aims to provide geometrically consistent observations across distortion regimes, especially in peripheral image regions where geometry-based detection is least reliable. Details of the architecture and training procedure are provided in Appendix~\ref{app:extractor_architecture}.

\vspace{-0.3cm}
\subsection{Coverage- and Observability-Aware Selector}
\label{sec:coverage_observability_selector}

The selector follows a simple principle from Sec.~\ref{sec:failure_analysis}: useful calibration frames should not only cover the image, but also provide stable and separable parameter-update directions during initialization. We therefore evaluate each candidate frame using two geometry-aware criteria: point-based projective isotropy and directed radial span. These criteria reject observations likely to induce unstable pose estimation or intrinsic-parameter coupling, while retaining partially visible targets when their valid corners remain informative.

Let \(\mathbf{u}_j\) denote a valid detected target corner, where validity requires finite image coordinates and sufficient detection confidence. The first criterion, \textit{projective isotropy}, measures whether the target view provides a stable pose-estimation proxy. Using valid correspondences between board-plane corners and image corners, we estimate a homography and compute its local projective Jacobian \(\mathbf{J}_{\mathrm{proj}}\) at the board center. The isotropy score is
{
\begin{equation}
s_{\mathrm{iso}}
=
\frac{\sigma_{\min}(\mathbf{J}_{\mathrm{proj}})}
{\sigma_{\max}(\mathbf{J}_{\mathrm{proj}})},
\end{equation}
}
where \(\sigma_{\min}\) and \(\sigma_{\max}\) are the smallest and largest singular values. A larger \(s_{\mathrm{iso}}\) indicates a less degenerate target projection and more stable pose initialization.

The second criterion, \textit{directed radial span}, measures whether a target observation sufficiently excites radial projection variation. Let \(\mathbf{o}=(W/2,H/2)^\top\) be the image center, and let \(R\) be the maximum valid radius from \(\mathbf{o}\) to the usable fisheye-mask boundary. For a frame, the primary radial direction is defined as
{
\begin{equation}
\mathbf{a}
=
\frac{\bar{\mathbf{u}}-\mathbf{o}}
{\|\bar{\mathbf{u}}-\mathbf{o}\|_2},
\qquad
\bar{\mathbf{u}}=\frac{1}{N}\sum_j \mathbf{u}_j ,
\end{equation}
}
with a principal-component direction used when \(\bar{\mathbf{u}}\) is too close to the image center. Each valid corner is projected onto this direction:
{
\begin{equation}
t_j
=
\frac{(\mathbf{u}_j-\mathbf{o})^\top\mathbf{a}}{R}.
\end{equation}
}
The directed radial span score is
{
\begin{equation}
s_{\mathrm{drs}}
=
\frac{\max_j t_j-\min_j t_j}{2}.
\end{equation}
}
This diameter-normalized signed span captures how broadly the target covers the radial direction, including cases where the board crosses the image center. A larger \(s_{\mathrm{drs}}\) helps separate focal scale from projection-shape parameters.

The selector organizes frames according to the calibration pipeline:
1) \textbf{Anchor selection:} choose frames that pass both projective-isotropy and directed-radial-span gates to guide intrinsic initialization with stable, decoupled observations.
2) \textbf{Co-visible selection:} select timestamp-matched frames that pass the per-camera gates in both cameras, providing reliable multi-camera constraints for stereo extrinsic estimation.
3) \textbf{Mono-fill:} add monocular frames to cover weakly observed image regions after Co-visible selection, without overwhelming initialization with low-quality observations.
Thus, CO-Calib constructs a frame sequence that guides the optimizer progressively beyond merely maximizing the number or uniformity of observations.

\begin{figure}[t]
\centering
\vspace{0.1cm}
\includegraphics[width= 0.9\columnwidth]{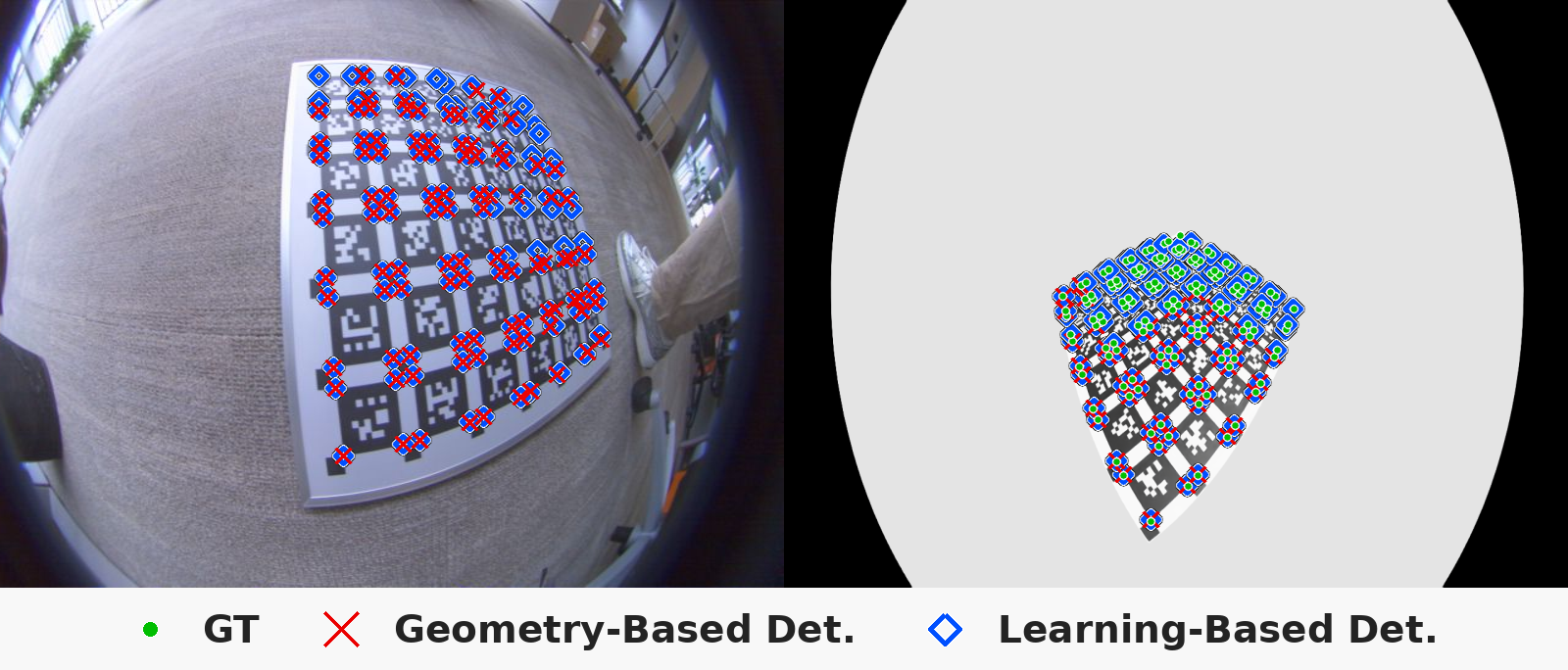}
\vspace{-0.4cm}
\caption{
Comparisons between the proposed learning-based detector and the geometry-based approach.
}
\label{fig:exp1-detector-improvement}
\vspace{-0.5cm}
\end{figure}

\vspace{-0.3cm}
\section{Experiments}
\label{sec:experiments}
\subsection{Experimental Setup}
\label{sec:experiements_setup}
To comprehensively evaluate the proposed calibration method, we collected two types of calibration data: synthetic data generated in simulation and real-world data captured using a multi-fisheye camera system. 

In the simulation experiments, we constructed 16 stereo-camera configurations by combining four FoVs, i.e., \(180^\circ\), \(200^\circ\), \(220^\circ\), and \(240^\circ\), with four relative yaw angles, i.e., \(0^\circ\), \(60^\circ\), \(90^\circ\), and \(120^\circ\). 
The stereo baseline was fixed to 5 cm. 
For each configuration, we generated 100 calibration sequences, each containing 480 frames at a resolution of \(1280 \times 720\). 
All cameras followed the \(f\)-theta projection model. 
The camera poses were randomly sampled in the space in front of the calibration board, with randomized viewing directions. 
The ground truth includes the camera poses and the image-plane locations of the calibration targets.

\begin{figure}[t]
\centering
\vspace{0.1cm}
\includegraphics[width= 0.95\columnwidth]{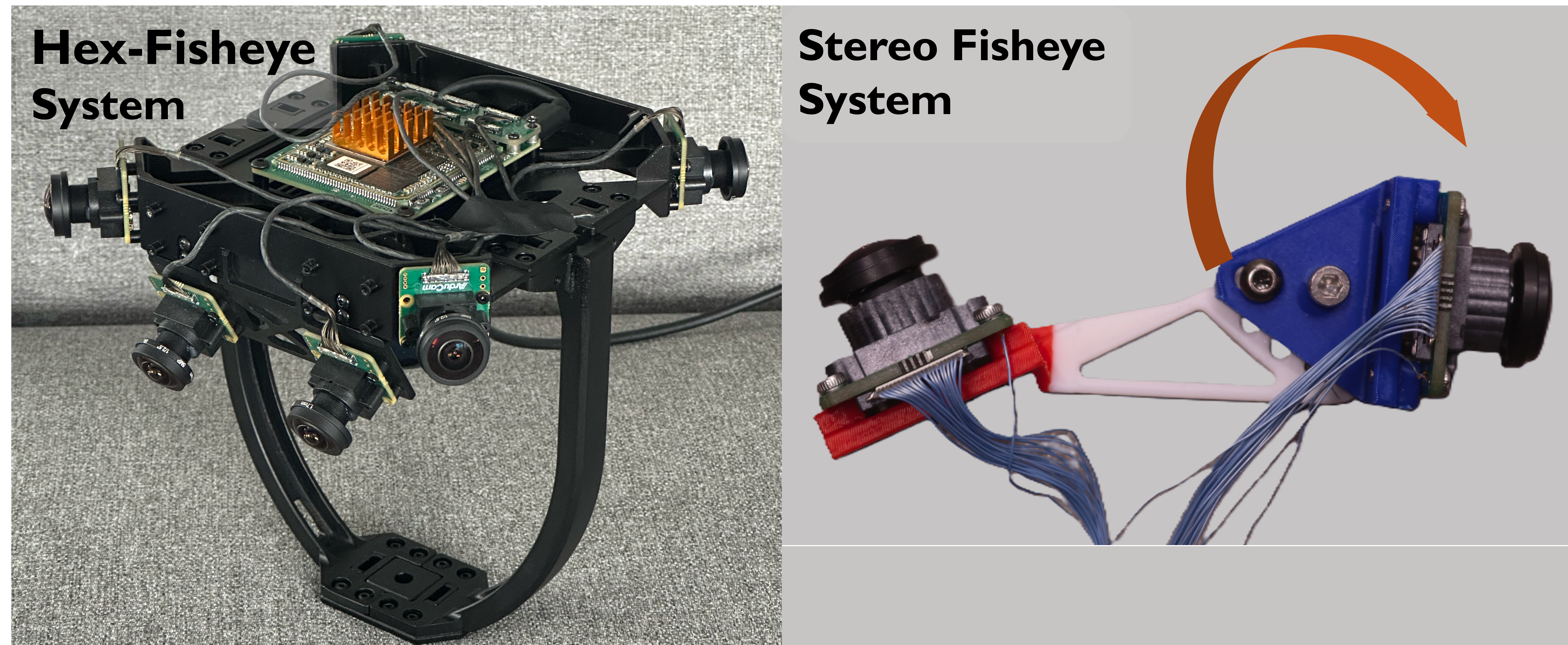}
\vspace{-0.3cm}
\caption{Visualization of real-world camera settings.}
\label{fig:hex-camera-system}
\vspace{-0.5cm}
\end{figure}

In the real-world experiments, calibration data were collected using the typical stereo fisheye camera pairs with relative yaw rotation of \(0^\circ\), \(30^\circ\), \(60^\circ\), \(90^\circ\), and \(120^\circ\), and a hexagonal fisheye camera system shown in Fig.~\ref{fig:hex-camera-system}. Five calibration sequences were recorded for each stereo configuration, and ten sequences were collected for the Hex-Fisheye camera.

For all experiments, the thresholds are set to \(s_{\mathrm{iso}}=0.3\) and \(s_{\mathrm{drs}}=110/\mathrm{FoV}\), where a rough FoV value can be obtained from the camera manufacturer in real-world applications. For the Co-visible selection and Mono-fill stages, both selection criteria are scaled by a factor of 0.6 to relax the constraints and increase the number of candidate frames.

\vspace{-0.5cm}
\subsection{Spatial Constraint Improvement}
We compare the proposed learning-based detector with geometry-based detection methods in terms of mean detection error and recall rate. The results in Table~\ref{tab:exp1_3_task12_subpixel_vs_kalibr} show that the proposed detector significantly improves the recall rate while also achieving a lower mean detection error, demonstrating its superior robustness and improved accuracy. The proposed detector also provides more reliable estimates of frame-level \(s_{\mathrm{iso}}\) and \(s_{\mathrm{drs}}\).

\begin{table}[H]
  \vspace{-0.5cm}
  \centering
  \caption{Detection recall and localization accuracy}
  \vspace{-0.3cm}
  \label{tab:exp1_3_task12_subpixel_vs_kalibr}
  \setlength{\tabcolsep}{2pt}
  \renewcommand{\arraystretch}{0.8}
  \scriptsize

  \begin{tabular*}{\columnwidth}{@{\extracolsep{\fill}}lcc}
  \toprule
  \textbf{FoV}
  & \shortstack{\textbf{Learning-based Det.}\\\textbf{Recall / Mean Error px}}
  & \shortstack{\textbf{Geometry-based Det.}\\\textbf{Recall / Mean Error px}} \\
  \midrule
  180 & 0.9330 / 0.7841 & 0.7085 / 0.8377 \\
  200 & 0.9344 / 0.7944 & 0.7049 / 0.8380 \\
  220 & 0.9255 / 0.7941 & 0.7077 / 0.8304 \\
  240 & 0.9263 / 0.8277 & 0.6838 / 0.8553 \\
  \bottomrule
  \end{tabular*}
  \vspace{-0.7cm}
\end{table}

\subsection{Calibration on Synthetic and Real-world Data}
\label{sec:calibration_experiment}

We first conduct a comprehensive comparison between CO-Calib and Kalibr on the synthetic benchmark. As shown in Table~\ref{tab:synthetic_results}, CO-Calib consistently improves the calibration success rate across different FoVs and stereo configurations. In particular, the overall success rate increases from 68.1\% with Kalibr to 99.3\% with CO-Calib. Meanwhile, CO-Calib also reduces the overall extrinsic calibration error from 0.54/0.029 to 0.18/0.021 in translation and rotation, respectively. For several configurations from FoV180 to FoV220, CO-Calib shows a slightly higher pose estimation error than Kalibr. This is mainly because the proposed selection strategy uses fewer frames for optimization, which may reduce the trajectory coverage available for camera pose estimation. Nevertheless, the selected frames provide better geometric constraints for intrinsic and extrinsic calibration, leading to substantially higher robustness and more accurate extrinsic estimates.

We then perform an ablation study on the proposed coverage- and observability-aware selector. For each selected sequence, we construct a same-size random subset and run the same calibration pipeline. The random subsets lead to a much lower overall success rate, dropping from 99.3\% to 30.9\%. This confirms that the improvement does not simply come from reducing the number of frames, but from selecting frames with better coverage and observability for calibration.

Finally, we evaluate the effect of removing the initialization stage by comparing the standard three-stage pipeline with a BA-only two-stage pipeline. Without initialization, the overall success rate decreases to 13.5\%, demonstrating that a reliable initialization remains critical for robust calibration, especially under wide-FoV and challenging stereo configurations. The detailed statistical results are reported in Table~\ref{tab:synthetic_results}.

We further conduct real-world experiments to evaluate the practical performance of different calibration methods. Since ground-truth camera parameters and camera poses are unavailable in real-world data, we use extrinsic consistency across repeated calibration sequences as a proxy metric. Specifically, lower standard deviations of the estimated baseline length and relative rotation angle indicate more stable calibration results. We compared our methods with Kalibr and Basalt\cite{basalt}. As shown in Table~\ref{tab:realworld_results}, CO-Calib achieves consistently high success rates across all stereo configurations. On the standard stereo configurations, CO-Calib obtains extrinsic consistency comparable to Kalibr, with translation differences typically within sub-millimeter range and rotation differences within a small fraction of a degree. More importantly, on the challenging Hex-Fisheye dataset, Kalibr fails on all sequences, while CO-Calib succeeds on all 10 trials and maintains high extrinsic consistency. These results demonstrate that CO-Calib provides robust and accurate calibration performance in both controlled synthetic settings and challenging real-world scenarios.

\begin{table*}[t]
\centering

\caption{Synthetic calibration results.
SR denotes success rate. ATE and Ext. denote pose and extrinsic errors, reported as \(t/r\) in mm/deg and averaged over strict-success trials. 
Bold marks the best SR.}
\vspace{-0.2cm}

\label{tab:synthetic_results}
\scriptsize
\setlength{\tabcolsep}{1.8pt}
\renewcommand{\arraystretch}{0.95}

\begin{tabular*}{\textwidth}{@{\extracolsep{\fill}}lcccccccccccc}
\toprule
\textbf{Method}
& \multicolumn{3}{c}{\textbf{FoV180\_0}}
& \multicolumn{3}{c}{\textbf{FoV180\_60}}
& \multicolumn{3}{c}{\textbf{FoV180\_90}}
& \multicolumn{3}{c}{\textbf{FoV180\_120}} \\
\cmidrule(lr){2-4}
\cmidrule(lr){5-7}
\cmidrule(lr){8-10}
\cmidrule(lr){11-13}
& \textbf{SR} & \textbf{ATE} & \textbf{Ext.}
& \textbf{SR} & \textbf{ATE} & \textbf{Ext.}
& \textbf{SR} & \textbf{ATE} & \textbf{Ext.}
& \textbf{SR} & \textbf{ATE} & \textbf{Ext.} \\
\midrule
Kalibr      & 97\% & 3.14/0.067 & 0.16/0.005 & \textbf{100\%} & 3.23/0.069 & 0.49/0.021 & \textbf{100\%} & 3.28/0.076 & 0.64/0.034 & \textbf{100\%} & 3.24/0.079 & 0.77/0.047 \\
CO-Calib    & \textbf{100\%} & 4.30/0.159 & 0.15/0.009 & \textbf{100\%} & 4.06/0.151 & 0.14/0.025 & 99\% & 4.16/0.146 & 0.25/0.018 & \textbf{100\%} & 4.83/0.189 & 0.40/0.086 \\
Random-subset & 39\% & 5.93/0.237 & 0.14/0.017 & 42\% & 3.99/0.148 & 0.20/0.035 & 35\% & 4.23/0.151 & 0.27/0.020 & 24\% & 4.63/0.170 & 0.53/0.052 \\
BA-only two-stage    & 29\% & 5.44/0.192 & 0.16/0.013 & 28\% & 4.77/0.204 & 0.40/0.041 & 17\% & 4.64/0.177 & 0.37/0.028 & 12\% & 5.33/0.212 & 0.61/0.098 \\

\midrule
\textbf{Method}
& \multicolumn{3}{c}{\textbf{FoV200\_0}}
& \multicolumn{3}{c}{\textbf{FoV200\_60}}
& \multicolumn{3}{c}{\textbf{FoV200\_90}}
& \multicolumn{3}{c}{\textbf{FoV200\_120}} \\
\cmidrule(lr){2-4}
\cmidrule(lr){5-7}
\cmidrule(lr){8-10}
\cmidrule(lr){11-13}
& \textbf{SR} & \textbf{ATE} & \textbf{Ext.}
& \textbf{SR} & \textbf{ATE} & \textbf{Ext.}
& \textbf{SR} & \textbf{ATE} & \textbf{Ext.}
& \textbf{SR} & \textbf{ATE} & \textbf{Ext.} \\
\midrule
Kalibr      & 97\% & 3.14/0.070 & 0.17/0.005 & 98\% & 3.24/0.071 & 0.53/0.019 & 97\% & 3.37/0.085 & 0.65/0.037 & 92\% & 3.35/0.088 & 0.79/0.047 \\
CO-Calib    & \textbf{100\%} & 3.14/0.078 & 0.17/0.006 & \textbf{99\%} & 3.24/0.085 & 0.10/0.017 & \textbf{100\%} & 3.57/0.105 & 0.11/0.021 & \textbf{100\%} & 4.31/0.152 & 0.39/0.051 \\
Random-subset & 59\% & 5.17/0.182 & 0.15/0.011 & 52\% & 4.24/0.164 & 0.21/0.026 & 54\% & 4.78/0.185 & 0.41/0.028 & 50\% & 4.85/0.185 & 0.50/0.073 \\
BA-only two-stage & 22\% & 5.17/0.185 & 0.14/0.013 & 26\% & 4.31/0.165 & 0.21/0.029 & 19\% & 4.84/0.194 & 0.39/0.032 & 20\% & 5.21/0.207 & 0.64/0.074 \\

\midrule
\textbf{Method}
& \multicolumn{3}{c}{\textbf{FoV220\_0}}
& \multicolumn{3}{c}{\textbf{FoV220\_60}}
& \multicolumn{3}{c}{\textbf{FoV220\_90}}
& \multicolumn{3}{c}{\textbf{FoV220\_120}} \\
\cmidrule(lr){2-4}
\cmidrule(lr){5-7}
\cmidrule(lr){8-10}
\cmidrule(lr){11-13}
& \textbf{SR} & \textbf{ATE} & \textbf{Ext.}
& \textbf{SR} & \textbf{ATE} & \textbf{Ext.}
& \textbf{SR} & \textbf{ATE} & \textbf{Ext.}
& \textbf{SR} & \textbf{ATE} & \textbf{Ext.} \\
\midrule
Kalibr      & 33\% & 3.25/0.081 & 0.18/0.005 & 64\% & 3.32/0.081 & 0.52/0.025 & 56\% & 3.60/0.108 & 0.69/0.052 & 54\% & 3.54/0.106 & 0.82/0.049 \\
CO-Calib    & \textbf{100\%} & 4.09/0.130 & 0.16/0.008 & \textbf{100\%} & 3.53/0.104 & 0.11/0.027 & \textbf{99\%} & 3.95/0.125 & 0.17/0.039 & \textbf{97\%} & 4.00/0.123 & 0.28/0.018 \\
Random-subset & 13\% & 4.70/0.163 & 0.15/0.013 & 33\% & 3.69/0.121 & 0.14/0.031 & 21\% & 4.53/0.175 & 0.37/0.051 & 20\% & 4.31/0.146 & 0.34/0.025 \\
BA-only two-stage   & 3\% & 4.64/0.151 & 0.15/0.010 & 10\% & 3.77/0.128 & 0.14/0.022 & 9\% & 4.70/0.176 & 0.40/0.054 & 8\% & 4.13/0.141 & 0.46/0.024 \\

\midrule
\textbf{Method}
& \multicolumn{3}{c}{\textbf{FoV240\_0}}
& \multicolumn{3}{c}{\textbf{FoV240\_60}}
& \multicolumn{3}{c}{\textbf{FoV240\_90}}
& \multicolumn{3}{c}{\textbf{FoV240\_120}} \\
\cmidrule(lr){2-4}
\cmidrule(lr){5-7}
\cmidrule(lr){8-10}
\cmidrule(lr){11-13}
& \textbf{SR} & \textbf{ATE} & \textbf{Ext.}
& \textbf{SR} & \textbf{ATE} & \textbf{Ext.}
& \textbf{SR} & \textbf{ATE} & \textbf{Ext.}
& \textbf{SR} & \textbf{ATE} & \textbf{Ext.} \\
\midrule
Kalibr      & 12\% & 3.56/0.099 & 0.24/0.006 & 41\% & 3.40/0.089 & 0.54/0.032 & 25\% & 3.37/0.099 & 0.64/0.078 & 23\% & 3.41/0.108 & 0.82/0.058 \\
CO-Calib    & \textbf{98\%} & 3.35/0.076 & 0.18/0.005 & \textbf{98\%} & 3.07/0.056 & 0.14/0.009 & \textbf{98\%} & 2.97/0.059 & 0.17/0.027 & \textbf{100\%} & 3.03/0.067 & 0.22/0.015 \\
Random-subset  & 12\% & 3.96/0.105 & 0.19/0.006 & 22\% & 3.07/0.061 & 0.12/0.009 & 13\% & 3.47/0.106 & 0.20/0.046 & 5\% & 3.80/0.141 & 0.41/0.035 \\
BA-only two-stage    & 1\% & 3.31/0.073 & 0.18/0.005 & 6\% & 2.99/0.060 & 0.20/0.008 & 5\% & 3.27/0.097 & 0.17/0.065 & 1\% & 3.88/0.138 & 0.15/0.023 \\

\bottomrule
\end{tabular*}
\vspace{-0.5cm}
\end{table*}

\begin{table}[t]
  \centering
  \caption{Overall synthetic calibration results across all 16 configurations.}
  \vspace{-0.2cm}
  \label{tab:synthetic_overall_results}
  \scriptsize
  \setlength{\tabcolsep}{4pt}
  \renewcommand{\arraystretch}{0.95}
  \begin{tabular*}{\columnwidth}{@{\extracolsep{\fill}}lccc}
    \toprule
    \textbf{Method} & \textbf{SR} & \textbf{ATE} & \textbf{Ext.} \\
    \midrule
    Kalibr            & 68.1\% & 3.27/0.079 & 0.54/0.029 \\
    CO-Calib          & \textbf{99.3\%} & 3.69/0.111 & \textbf{0.18/0.021} \\
    Random-subset     & 30.9\% & 4.43/0.160 & 0.26/0.029 \\
    BA-only two-stage & 13.5\% & 4.74/0.178 & 0.31/0.033 \\
    \bottomrule
  \end{tabular*}
  \vspace{-0.5cm}
\end{table}

\begin{table*}[t]
\centering
\caption{Real-world extrinsic consistency across camera configurations. 
Ext. Std. denotes the standard deviation of estimated extrinsics across repeated trials, reported as \(t/r\) in mm/deg, where \(t\) is baseline-length variation and \(r\) is relative-rotation variation. 
SR is reported as successful trials over total trials.}
\vspace{-0.2cm}
\label{tab:realworld_results}
\scriptsize
\setlength{\tabcolsep}{1.8pt}
\renewcommand{\arraystretch}{0.95}

\begin{tabular*}{\textwidth}{@{\extracolsep{\fill}}l*{12}{c}}
\toprule
\textbf{Method}
& \multicolumn{2}{c}{\textbf{Rot-0}}
& \multicolumn{2}{c}{\textbf{Rot-30}}
& \multicolumn{2}{c}{\textbf{Rot-60}}
& \multicolumn{2}{c}{\textbf{Rot-90}}
& \multicolumn{2}{c}{\textbf{Rot-120}}
& \multicolumn{2}{c}{\textbf{Hex-Fisheye}} \\
\cmidrule(lr){2-3}
\cmidrule(lr){4-5}
\cmidrule(lr){6-7}
\cmidrule(lr){8-9}
\cmidrule(lr){10-11}
\cmidrule(lr){12-13}
&
\textbf{Ext. Std.} & \textbf{SR}
& \textbf{Ext. Std.} & \textbf{SR}
& \textbf{Ext. Std.} & \textbf{SR}
& \textbf{Ext. Std.} & \textbf{SR}
& \textbf{Ext. Std.} & \textbf{SR}
& \textbf{Ext. Std.} & \textbf{SR} \\

\midrule
Kalibr
& 0.16/0.142 & 5/5
& 0.08/0.256 & 5/5
& 0.17/0.388 & 5/5
& 0.71/0.653 & 5/5
& 0.64/0.118 & 5/5
& -- & 0/10 \\
CO-Calib
& 0.21/0.146 & 5/5
& 0.14/0.247 & 5/5
& 0.33/0.357 & 5/5
& 0.43/0.649 & 5/5
& 0.47/0.267 & 5/5
& 0.97/0.048 & 10/10 \\
Basalt
& 1.57/0.456 & 5/5
& 5.82/0.279 & 4/5
& 4.39/0.632 & 5/5
& 14.20/1.088 & 5/5
& 11.30/1.250 & 5/5
& 51.23/6.626 & 4/10 \\
\bottomrule
\end{tabular*}
\vspace{-0.5cm}
\end{table*}

\vspace{-0.3cm}
\section{Conclusion}
This paper studies why standard calibration pipelines perform unstably in multi-fisheye systems. 
Through controlled failure-oriented analysis, we demonstrate that calibration failures cannot be fully explained by detector recall loss or by global spatial constraint distribution. 
Instead, the key failure mode occurs during intrinsic initialization: observations with limited radial span can couple focal scale with fisheye projection-shape parameters, leading to ill-conditioned local updates. 
This finding highlights the central message of this work: observation quality should be measured not only by coverage or sample count, but also by whether observations provide separable and stable parameter-update directions.

Based on this analysis, we propose CO-Calib, a plug-in framework that combines a learning-based calibration-target detector with an error-analysis-guided frame selector. 
The selector constructs calibration data through anchor, Co-visible selection, and Mono-fill stages, allowing selected observations to better guide intrinsic initialization and subsequent multi-fisheye optimization. 
Across comprehensive simulations and real captures, the constructed observation sets make calibration substantially more reliable, raising synthetic success from 68.1\% to 99.3\% with lower extrinsic error and yielding more consistent real-world extrinsics on challenging rigs such as Hex-Fisheye.



\vspace{-0.4cm}
\bibliography{references}

\vspace{-0.3cm}
\appendix
\subsection{Spatial Distribution Metrics}
\label{app:spatial_distribution_metrics}
This section details the metrics used in Table~\ref{tab:distribution_success_failure_gt_raw_nohuber}. For each setting and camera, the image plane is divided into regular grid cells. Let \(C_{m,i}\) denote the number of valid observation points in cell \(i\) for case \(m\). Given the successful set \(\mathcal{S}\) and failed set \(\mathcal{F}\), with sizes \(n_S=|\mathcal{S}|\) and \(n_F=|\mathcal{F}|\), we compute the average count maps,
{
\begin{equation}
S_i
=
\frac{1}{n_S}
\sum_{m\in\mathcal{S}}
C_{m,i},
\qquad
F_i
=
\frac{1}{n_F}
\sum_{m\in\mathcal{F}}
C_{m,i}.
\end{equation}
}

Let \(N_S=\sum_i S_i\) and \(N_F=\sum_i F_i\) be the average total numbers of observations, and define \(P_i=S_i/N_S\) and \(Q_i=F_i/N_F\) as the corresponding normalized spatial distributions:
{
\begin{equation}
D_{\mathrm{qty}}
=
\frac{|N_S-N_F|}{N_S+N_F},
\qquad
D_{\mathrm{sp}}
=
\frac{1}{2}
\sum_i |P_i-Q_i|.
\end{equation}
}
Here, \(D_{\mathrm{qty}}\) measures only the total observation-count difference, while \(D_{\mathrm{sp}}\) measures the normalized image-space distribution difference after removing the effect of total count.

To estimate the distance expected from random grouping, we repeatedly permute the success/failure labels within the same setting while preserving the original group sizes \(n_S\) and \(n_F\). For each random relabeling \(r\), we recompute the same spatial distance \(D_{\mathrm{sp}}^{(r)}\) and define
$
D_{\mathrm{rand}}
=
\mathrm{median}_r\, D_{\mathrm{sp}}^{(r)}$.

\vspace{-0.5cm}
\subsection{Implementation Details of Target Detector}
\label{app:extractor_architecture}
\subsubsection{Network Architecture}

\begin{figure}[t]
\centering
\includegraphics[width= 0.8\columnwidth]{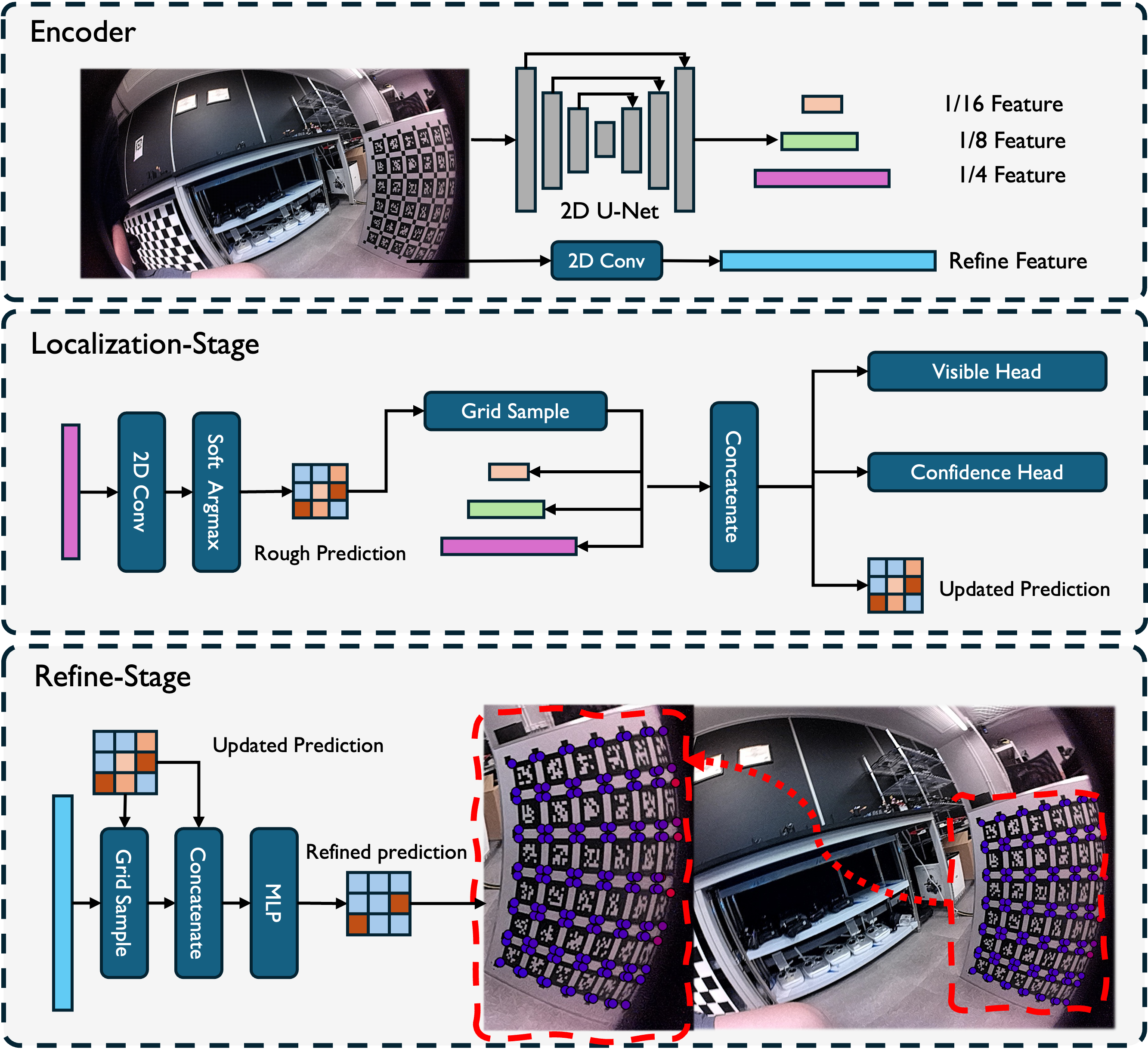}
\vspace{-0.3cm}
\caption{Architecture of the proposed calibration-target detector. 1) Encoder provides multi-scale features. 2) The localization stage produces rough and updated corner predictions with visibility, learnability, and confidence estimates. 3) The refinement stage outputs final corner locations.}
\vspace{-0.6cm}
\label{fig:nn-structure}
\end{figure}

As shown in Fig.~\ref{fig:nn-structure}, the detector adopts a multi-stage architecture.
A U-Net-style encoder extracts multi-level features, while a shallow side branch keeps original-resolution details.
The network first generates rough corner predictions from the 1/4 heatmap and updates them with sampled 1/8 and 1/16 features, producing updated predictions. 
Additionally, several auxiliary supervision signals are introduced at this stage to improve visibility estimation, prediction reliability, and cross-slot identity consistency. 
Finally, the shallow side branch further refines the updated predictions and outputs the final refined predictions.

\subsubsection{Online Physically-Grounded Data Generation}

Collecting diverse, large-scale real calibration data is costly and challenging across the wide range of lenses, fields of view, and distortion patterns encountered in multi-fisheye systems. 
To address this, we train the detector using an online physically grounded data generation pipeline.

Each iteration instantiates a target in canonical 3D coordinate space and randomly transforms it to simulate diverse poses. 
We then render them onto the image plane by a physically consistent projection process with randomized camera parameters to produce target appearances under various imaging characteristics. 
The rendered target is further composited with varying backgrounds, while ground-truth corner locations, visibility labels, and auxiliary supervision labels are generated simultaneously.
This online strategy offers scalable supervision with broad geometric diversity and improves detector robustness across various multi-fisheye setups.

\subsubsection{Training Recipe}
We optimize a multi-task objective for coarse-to-fine localization and reliability.
It supervises heatmaps; rough, updated, and refined coordinates; visibility, learnability, and confidence; plus exclusion to prevent corner channels from firing at the same location:
{
\begin{equation}
\begin{aligned}
\mathcal{L} =\;&
\lambda_{h+}\mathcal{L}_{h+}
+ \lambda_{h-}\mathcal{L}_{h-}
+ \lambda_{e}\mathcal{L}_{e} \\
&+ \sum_{s \in \{r,u,f\}} \lambda_s \mathcal{L}_s
+ \sum_{a \in \{v,l,c\}} \lambda_a \mathcal{L}_a ,
\end{aligned}
\end{equation}
}
where $r$, $u$, and $f$ denote the rough, updated, and refined coordinate stages, respectively, and $v$, $l$, and $c$ denote visibility, learnability, and confidence estimation.

All coordinate stages share a weighted regression loss:
{
\begin{equation}
\mathcal{L}_s
=
\frac{1}{\sum_i w_i^{(s)}}
\sum_i
w_i^{(s)} \,
\rho_s(\hat{p}_i^{(s)}, p_i),
\quad s \in \{r,u,f\},
\end{equation}
}
where $p_i$ and $\hat{p}_i^{(s)}$ are ground-truth and predicted corner-$i$ locations, and $\rho_s$ is the stage-specific loss. The rough stage uses $w_i^{(r)} = w_i^{\mathrm{sup}}$, while updated and refined stages use $w_i^{(u)} = w_i^{(f)} = m_i^{\mathrm{vis}}$, with $m_i^{\mathrm{vis}} \in \{0,1\}$ and $w_i^{\mathrm{sup}} \in [0,1]$ determined by visibility and learnability.
Heatmap supervision consists of positive and negative terms:
{
\begin{equation}
\mathcal{L}_{h\pm}
=
\frac{1}{\sum_i \beta_i^{\pm}}
\sum_i
\beta_i^{\pm}\,
\mathrm{BCE}(H_i, T_i^{\pm}),
\end{equation}
}
where $H_i$ is the predicted heatmap, $\beta_i^{+}=w_i^{\mathrm{sup}}$, $T_i^{+}=G_i$, $\beta_i^{-}=1-m_i^{\mathrm{vis}}$, $T_i^{-}=0$, and BCE spans all spatial locations.

To reduce cross-slot confusion, exclusion penalizes channel $k$ at other visible corners:
{
\begin{equation}
\mathcal{L}_{e}
=
\frac{1}{Z}
\sum_{j \neq k}
m_j^{\mathrm{vis}} m_k^{\mathrm{vis}}
\,
\mathrm{BCE}(H_k(p_j), 0),
\end{equation}
}
where $H_k(p_j)$ denotes the response of channel $k$ sampled at the ground-truth location of corner $j$, and $Z$ is a normalization term.
The auxiliary heads use a shared weighted-loss form:
{
\begin{equation}
\mathcal{L}_a
=
\frac{1}{\sum_i \alpha_i^{(a)}}
\sum_i
\alpha_i^{(a)}\,
\phi_a(\hat{y}_i^{(a)}, y_i^{(a)}),
\quad a \in \{v,l,c\},
\end{equation}
}
where $\phi_v=\phi_c=\mathrm{BCE}(\cdot,\cdot)$, $\phi_l=\rho_l$, $\alpha_i^{(v)}=\alpha_i^{(c)}=1$, and $\alpha_i^{(l)}=m_i^{\mathrm{vis}}$. The confidence target is defined from the refined localization error as
{
\begin{equation}
y_i^{(c)}
=
m_i^{\mathrm{vis}}
\exp\!\left(
-\frac{\|\hat{p}_i^{(f)}-p_i\|_2^2}{2\sigma^2}
\right).
\end{equation}
}


\end{document}